\documentclass{article}

\usepackage{arxiv}

\usepackage[utf8]{inputenc} 
\usepackage[T1]{fontenc}    
\usepackage[colorlinks = true,
            linkcolor = green,
            urlcolor  = green,
            citecolor = blue,
            anchorcolor = blue]{hyperref}
\usepackage{url}            
\usepackage{booktabs}       
\usepackage{amsfonts}       
\usepackage{nicefrac}       
\usepackage{microtype}      
\usepackage{graphicx}
\usepackage{multirow}
\usepackage{lipsum}
\newcommand\blfootnote[1]{%
  \begingroup
  \renewcommand\thefootnote{}\footnote{#1}%
  \addtocounter{footnote}{-1}%
  \endgroup
}
            
\title{Analysing Risk of Coronary Heart Disease through Discriminative Neural Networks}
\author{Ayush Khaneja\textsuperscript{\textdagger}, Siddharth Srivastava\textsuperscript{\textasteriskcentered}\textsuperscript{\textdaggerdbl}, Astha Rai\textsuperscript{\textasteriskcentered}, A S Cheema\textsuperscript{\textasteriskcentered}, P K Srivastava\textsuperscript{\textasteriskcentered}\\
\textsuperscript{\textdagger}Vellore Institute of Technology, Vellore, India \\ \textsuperscript{\textasteriskcentered}Centre for Development of Advanced Computing, Noida, India\\
Email: ayush.khaneja@gmail.com, \{siddharthsrivastava, asthar, ascheema, pksrivatava\}@cdac.in}

\begin{document}
\maketitle
\begin{abstract}
The application of data mining, machine learning and artificial intelligence techniques in the field of diagnostics is not a new concept, and these techniques have been very successfully applied in a variety of applications, especially in dermatology and cancer research. But, in the case of medical problems that involve tests resulting in true or false (binary classification), the data generally has a class imbalance with samples majorly belonging to one class (ex: a patient undergoes a regular test and the results are false). Such disparity in data causes problems when trying to model predictive systems on the data. In critical applications like diagnostics, this class imbalance cannot be overlooked and must be given extra attention. In our research, we depict how we can handle this class imbalance through neural networks using a discriminative model and contrastive loss using a Siamese neural network structure. Such a model does not work on a probability-based approach to classify samples into labels. Instead it uses a distance-based approach to differentiate between samples classified under different labels. The code is available at \url{https://tinyurl.com/DiscriminativeCHD}
\end{abstract}


\section{Introduction}
\label{sec: intro}
\blfootnote{During this work, Ayush Khaneja was an intern at Centre for Development of Advanced Computing, Noida, India.  \textsuperscript{\textdaggerdbl}Corresponding Author}
The medical practitioners especially in tertiary care hospitals work under tremendous workload resulting in a considerable impact on patient care \cite{michtalik2013impact,lemaire2019stakeholder}. Further, machine learning has shown promise in assisting medical practitioners such as predicting diseases\cite{liu2018deep}, prognosis\cite{sun2019multimodal}, complications\cite{meyer2018machine} and many more \cite{beam2018big}. Therefore, there is a need for automated diagnostic systems that can work based on patient details such as age, gender, medical history etc. Also, the level of skill and knowledge of medical professionals is highly variable \cite{cuthbert1999expert,basugi2011comparison}. Through our work, we aim at providing a solution to these problems by developing a model that can perform binary classification (predicting if a patient is at risk of suffering from a particular disease or not) while handling problems of real-life data like class imbalance \cite{basugi2011comparison}. To demonstrate the effectiveness of the proposed method we select the  Heart diseases are one of the most common ailments in the current times due to a lifestyle of fast food, stress, no exercise, and vices like smoking and drinking. Thus, it becomes an important medical problem to be able to predict heart-related problems efficiently. We build our model to predict the risk of coronary heart disease in patients using the publicly available Framingham Heart Study  data\cite{kannel1971serum}. The data is captured over a wide variety of patients and is amongst the most established datasets for heart disease prediction \cite{zhu2018causal}. 

Automated prediction of diagnoses based on patient data has been a subject of interest for long now \cite{chen2017disease}. Various techniques have been successfully applied to different types of data and varied styles of diagnoses. Machine learning models made to classify data follow one of the two paradigms, generative or discriminative. Generative classifiers model how data is generated in order to classify a signal. Discriminative classifiers do not care how data is generated, they just categorize a given signal through a distance or similarity metric. In our work, we propose a discriminative approach to classification through the use of Siamese networks with a contrastive loss function for binary classification of coronary heart disease. We show that the proposed method is highly stable and handles bias due to class imbalance better than a vanilla neural network. 

The rest of the paper is organized as follows. In Section \ref{sec:rel}, we describe the works tackling health care data for disease classification. In Section \ref{sec:Approach}, we provide background of the techniques used and describe the proposed method. In Section \ref{sec:exp}. Finally, the conclusion is provided in Section \ref{sec:conc}.

\section{Related Works}\label{sec:rel}
A lot of research has been done in the field of disease prediction. But none of the system
is put in real life picture to the best of our knowledge. Some of the earliest Machine Learning based disease prediction was done by W. Nick et al in 1995\cite{street1995inductive}. They used a linear programming function for the prediction of the disease. Haq et al \cite{haq2018hybrid}  used Logistic regression to detect heart related problems from patient’s demographic and diagnosis data. Miotto et al \cite{miotto2016deep} used Decision Tree, Artificial Neural Network and Support Vector Machine for detecting Breast Cancer from demographic and EHR data of the patient. Benjamin Shickel et al \cite{shickel2017deep} conducted a survey on the deep learning technologies for the analysis of Electronic Health Record data. They compared in total of 6 projects in their paper. 

Guo et al \cite{guo2016robust} have studied time series with Recurrent Neural Network. In the WG-Learning, they introduce the weighted gradient to the online SGD for the RNN models, based on the local features of time series. The method enables to update the RNN models with down weighted gradients for outliers while full gradients for change points. Shiyue Zhang et al \cite{zhang2017medical} proposed a Variational Recurrent Neural Networks (VRNN) and Discriminative Neural Network for the analysis of lab test and their relation to diseases prediction. They used a dataset of more than 46000 patients covering 50 lab tests to diagnose 50 common diseases.
Detrano et al. Authors in \cite{detrano1989international} proposed a logistic regression classifier-based decision support system for heart disease classification. Leveraging large historical data in electronic health record (EHR), Edward Choi\cite{choi2016doctor} developed Doctor AI, a generic predictive model that covers observed medical conditions and medication uses. Doctor AI is a temporal model using recurrent neural networks (RNN) and was developed and applied to longitudinal time stamped EHR data from 260K patients and 2,128 physicians over 8 years. 

Edward Choi \cite{choi2016using} explored whether use of deep learning to model temporal relations among events in electronic health records (EHRs) would improve model performance in predicting initial diagnosis of heart failure (HF) compared to conventional methods that ignore temporality. Recurrent neural network (RNN) models using gated recurrent units (GRUs) were adapted to detect relations among time-stamped events (eg, disease diagnosis, medication orders, procedure orders, etc.) with a 12- to 18-month observation window of cases and controls. Model performance metrics were compared to regularized logistic regression, neural network, support vector machine, and K-nearest neighbor classifier approaches. In contrast to the above method, the proposed method works on the premise that it is important to learn a discriminative embedding among the classes to be classified rather than only a \textit{good} embedding. 

\section{Methodology}\label{sec:Approach}
\subsection{Background}
Prior to explaining the proposed method, we briefly describe the background concepts for the paper.

\subsubsection{Discriminative Networks}
Discriminative or conditional models are commonly used in supervised learning applications. Instead of modelling class-conditional probability distributions and prior probabilities, they directly estimate posterior probabilities directly from training data and do not try to model the core probability distributions. Essentially, discriminative models do not try to model how a particular data sample might have been generated. Instead, they measure how similar or dissimilar a data sample is from data of a particular class. Discriminative models do not learn how to predict the probability of data belonging to a particular class, but learn how to differentiate between data from different classes by observing its features.

\subsubsection{Siamese Networks}
They are a type of artificial neural networks that are made up of two or more identical networks that have identical weights. Working in tandem, each of these inner networks receives an input vector based on which an output vector is generated. These output vectors can then be compared to see how similar they are. Usually, Siamese networks are used to perform binary classification. This classification is done by seeing how similar or dissimilar a data sample is to samples of the two classes. The class exhibiting more similarity is decided as the predicted class. These networks generally find use in applications like facial identification, signature verification etc.

\subsubsection{Contrastive Loss}
It is a loss function typically used to learn the data’s discriminative features, i.e. learn to differentiate between samples from different classes. It is a distance-based loss function that tries to ensure that data samples that are semantically similar are embedded closer together, hence, calculated on pairs.	

\subsection{Proposed Network}
The network we propose has two identical feed-forward neural networks that work together and share weights. An input pair is given to the Siamese network and the inner networks receive one sample each from the pair. The inner networks then calculate the embeddings for their respective inputs and then the distance is calculated on these embeddings. This distance is then used to calculate the contrastive loss. Based on this calculate loss and the accuracy of the whole network, the weights are accordingly updated.

\begin{figure*}
    \centering
    \includegraphics[scale=0.6]{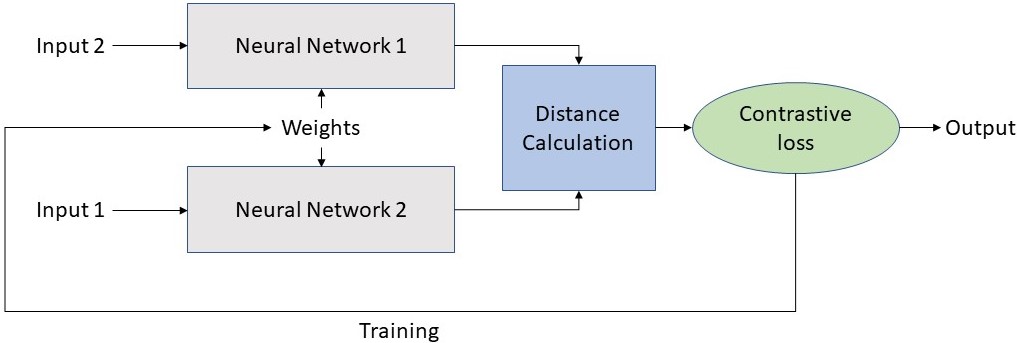}
    \caption{Structure of the proposed Siamese network}
    \label{fig:fig1}
\end{figure*}

\subsubsection{Missing value imputation}
All missing values were imputed and the whole data was normalized. Then, the input pairs were generated. To generate these pairs, we used random number generation. The first step was to split our data based on its label. Next, these new datasets were shuffled to ensure randomness. Then, pairs were created based on the generation of random numbers. random samples were picked from our new data sets and paired together. 100000 pairs were generated where data was picked from different classes. 50000 samples each were generated here data was selected from the same classes, i.e. 0 paired with 0 and 1 paired with 1. The 200000 pairs obtained were split in an 80-20 ratio to generate our training and testing data respectively. Since our goal is to differentiate between the two classes, we do not need to worry about samples being repeated in the data.

\section{Experiments}\label{sec:exp}
\subsection{Dataset} The dataset used is the Framingham dataset, which contains $4240$ samples and has $16$ columns. the properties of the dataset is given in Table $1$.

\begin{table}[]\label{tab:data}
\centering
\caption{Summary of the dataset}
\begin{tabular}{|p{15mm}|p{23mm}|p{8mm}|}
\hline
\multicolumn{1}{|c|}{\textbf{Data Type}} & \multicolumn{1}{c|}{\textbf{Column Name}} & \multicolumn{1}{c|}{\textbf{Missing Values}} \\ \hline
\textbf{Nominal}                         & male                                      & 0                                            \\ \hline
                                         & currentSmoker                             & 0                                            \\ \hline
                                         & BPMeds                                    & 53                                           \\ \hline
                                         & prevalentStroke                           & 0                                            \\ \hline
                                         & prevalentHyp                              & 0                                            \\ \hline
                                         & diabetes                                  & 0                                            \\ \hline
                                         & TenYearCHD (Label)                        & 0                                            \\ \hline
\textbf{Continuous}                      & age                                       & 0                                            \\ \hline
                                         & cigsPerDay                                & 29                                           \\ \hline
                                         & totChol                                   & 50                                           \\ \hline
                                         & sysBP                                     & 0                                            \\ \hline
                                         & diaBP                                     & 0                                            \\ \hline
                                         & BMI                                       & 19                                           \\ \hline
                                         & heartRate                                 & 1                                            \\ \hline
                                         & glucose                                   & 388                                          \\ \hline
\textbf{Discrete}                        & education                                 & 105                                          \\ \hline
\end{tabular}
\end{table}

\subsection{Base Network}
A simple feed-forward network with fully connected layers was trained on regular data and its characteristics were evaluated. This gave us a basis to improve upon. The Siamese network’s job was to improve upon the performance of this network. The following table summarizes the final results obtained on the base network after tweaking its hyperparameters. During training 25\% of the data was used for validation. The parameters of base network are shown in Table 2 along with training loss and confusion matrices over the classes. The goal was not only to increase accuracy, but to keep the number of ‘false negative’ errors (1 classified as 0) at a minimum, because in the case of medical data, a positive diagnosis being identified correctly is more important.

\begin{table}[h]
\centering
\caption{Parameter and experiments on base network}
\begin{tabular}{|l|l|}
\hline
\multirow{10}{*}{\textbf{Model Layers}}    & Input layer (size = 15)                                                                          \\ \cline{2-2} 
                                           & Hidden layer 1 (size = 256, L2 activity regularization = 0.01)                                   \\ \cline{2-2} 
                                           & Dropout = 0.175                                                                                  \\ \cline{2-2} 
                                           & Activation layer 1 (ReLU)                                                                        \\ \cline{2-2} 
                                           & Hidden layer 2 (size = 256, L2 activity regularization = 0.01)                                   \\ \cline{2-2} 
                                           & Dropout = 0.175                                                                                  \\ \cline{2-2} 
                                           & Activation layer 2 (ReLU)                                                                        \\ \cline{2-2} 
                                           & Output layer (size = 1, L2 activity regularization = 0.01)                                       \\ \cline{2-2} 
                                           & Dropout = 0.175                                                                                  \\ \cline{2-2} 
                                           & Activation layer 3 (Sigmoid)                                                                     \\ \hline
\multirow{2}{*}{\textbf{Optimizer}}        & Adam optimizer                                                                                   \\ \cline{2-2} 
                                           & (learning rate = 0.001)                                                                          \\ \hline
\textbf{Loss Function}                     & Binary Crossentropy                                                                              \\ \hline
\textbf{Evaluation Metric}                 & Accuracy                                                                                         \\ \hline
\textbf{Batch Size / No. of Epochs}        & 16 / 250                                                                                         \\ \hline
\multirow{2}{*}{\textbf{Class Weights}}    & Class 0: 1.0                                                                                     \\ \cline{2-2} 
                                           & Class 1: 5.0                                                                                     \\ \hline
\textbf{Loss on Training data}             & 1.01                                                                                             \\ \hline

\multirow{2}{*}{\textbf{Confusion Matrix}} & \multirow{2}{*}{\begin{tabular}[c]{@{}l@{}}{[} {[}508, 211{]},\\ {[} 42, 87{]} {]}\end{tabular}} \\
                                           &                                                                                                                                  \\ \hline
\end{tabular}
\end{table}

\subsection{Results on the proposed network}
The model was trained and its hyperparameters were tweaked to reach accuracies of almost 100\% in being able to differentiate between samples belonging to different classes. Again, during training, a 25\% validation split was used. Now, to make predictions, all that is needed is to take the new data and pair it with data from both the class 0 and the class 1. The network will tell us which class our data is closer to in terms of distance and we can the classify our new data accordingly. Table 2 summarizes the network parameters. Additionally, Fig 2 and 3 show the training and testing curves for accuracy and loss of the proposed network. 

\subsection{Comparison among networks} 
The results on the base network and proposed network are shown in Table 4. It can be observed that the proposed method obtains a significant improvement over the base network while achieving nearly 100\% accuracy. It shows that while classification based on feed-forward network work well if the data is balanced, the discriminative network, with their ability to learn a distance metric to map the input classes to separate embedding space, allow for stronger representation of the inputs, and hence result in more stable classification. This is indicated by the fact that while the precision for class 1 using the base network is 0.29 as compared to 0.92 for class 0, with the proposed network the precision is nearly same for both the classes.

\begin{table}[h]
\caption{Parameter and experiments on the proposed network}
\centering
\begin{tabular}{|l|l|}
\hline
\multirow{6}{*}{\textbf{Model Layers}}     & Input layer (Dense, size = 15),                       \\ \cline{2-2} 
                                           & Hidden layer 1 (Dense, size = 256, activation = ReLU) \\ \cline{2-2} 
                                           & Dropout = 0.2                                         \\ \cline{2-2} 
                                           & Hidden layer 2 (Dense, size = 256, activation = ReLU) \\ \cline{2-2} 
                                           & Dropout = 0.2                                         \\ \cline{2-2} 
                                           & Output layer (Dense, size = 256, activation = ReLU)   \\ \hline
\textbf{Optimizer}                         & RMSProp (learning rate = 0.001)                       \\ \hline
\textbf{Loss Function}                     & Contrastive Loss                                      \\ \hline
\textbf{Evaluation Metric}                 & Accuracy                                              \\ \hline
\textbf{Batch Size / Number of Epochs}     & 64 / 10                                               \\ \hline
\textbf{Loss on Training data}             & 0.0029                                                                      \\ \hline
\multirow{2}{*}{\textbf{Confusion Matrix}} & {[}{[}19913, 133{]},                                  \\ \cline{2-2} 
                                           & {[}0, 19954{]}{]}                                                                      \\ \hline
\end{tabular}
\end{table}

\begin{figure}
    \centering
    \includegraphics[scale=1.0]{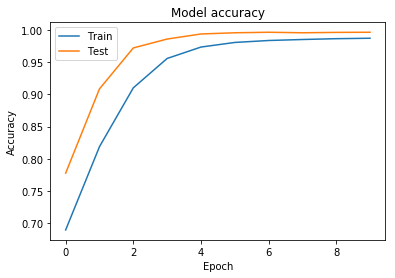}
    \caption{Accuracy}
    \label{fig:fig2}
\end{figure}
\begin{figure}
    \centering
    \includegraphics[scale=1.0]{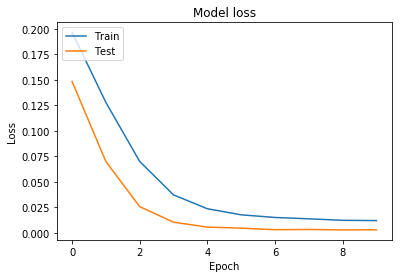}
    \caption{Loss}
    \label{fig:fig3}
\end{figure}

\begin{table}[]
\centering
\caption{Empirical Results on the Framingham dataset}
\begin{tabular}{|l|l|c|l|c|l|c|}
\hline
\multirow{2}{*}{\textbf{Method/metric}} & \multicolumn{2}{l|}{\textbf{Accuracy}}                           & \multicolumn{2}{l|}{\textbf{Precision}}                           & \multicolumn{2}{l|}{\textbf{Recall}}                              \\ \cline{2-7} 
                                        & \textbf{Training}           & \multicolumn{1}{l|}{\textbf{Test}} & \textbf{Class 0}          & \multicolumn{1}{l|}{\textbf{Class 1}} & \textbf{Class 0}          & \multicolumn{1}{l|}{\textbf{Class 1}} \\ \hline
\multicolumn{1}{|c|}{Base Network}      & \multicolumn{1}{c|}{0.72}   & 0.70                               & \multicolumn{1}{c|}{0.92} & 0.29                                  & \multicolumn{1}{c|}{0.71} & 0.67                                  \\ \hline
\multicolumn{1}{|c|}{Proposed Network}  & \multicolumn{1}{c|}{0.9966} & 0.9966                             & \multicolumn{1}{c|}{1.00} & 0.99                                  & \multicolumn{1}{c|}{0.99} & 1.00                                  \\ \hline
\end{tabular}
\end{table}

\section{Conclusion}\label{sec:conc}
In this paper we proposed a method based on discriminative network for classifying coronary heart disease. We showed that such networks are more stable on imbalanced data as compared to vanilla neural networks. In future, we believe that our work has potential to be integrated in practice with  Hospital Management Information Systems and used in real time for patient diagnosis.

\bibliographystyle{unsrt}  
\bibliography{references}  


\end{document}